\documentclass[lettersize,journal]{IEEEtran}
 
\usepackage{cite}
\usepackage[utf8]{inputenc} % allow utf-8 input
\usepackage[T1]{fontenc}    % use 8-bit T1 fonts
\usepackage{url}            % simple URL typesetting
\usepackage{booktabs}       % professional-quality tables
\usepackage{amsfonts}       % blackboard math symbols
\usepackage{nicefrac}       % compact symbols for 1/2, etc.
\usepackage{microtype}      % microtypography
\usepackage{enumitem}
\usepackage{amsmath}
\usepackage{amssymb}
\usepackage{multirow}
\usepackage{color}
\usepackage{xcolor}
\usepackage{colortbl}
\usepackage{subcaption}
\usepackage{pifont}
\usepackage[most]{tcolorbox} 
\tcbuselibrary{listingsutf8}
\usepackage{CJK}

\definecolor{cvprblue}{rgb}{0.21,0.49,0.74}
\definecolor{prompt}{rgb}{0.21,0.49,0.74}
\usepackage[pagebackref,breaklinks,colorlinks]{hyperref}

\newcommand{\dummy}[1]{}

\begin{document}
\title{Mono-InternVL-1.5:  Towards Cheaper and Faster Monolithic Multimodal Large Language Models }

\author{
    % \normalsize{}
     Gen Luo, Wenhan Dou, Wenhao Li, Zhaokai Wang, Xue Yang, Changyao Tian, Hao Li, \\Weiyun Wang, Wenhai Wang, Xizhou Zhu, Yu Qiao, Jifeng Dai$^\dagger$
     \thanks{
Gen Luo, Wenhao Li, Weiyun Wang and Yu Qiao are with Shanghai Artificial Intelligence Laboratory. 
Wenhan Dou, Xizhou Zhu and Jifeng Dai are with Tsinghua University. 
Changyao Tian, Hao Li and  Wenhai Wang are with The Chinese University of Hong Kong. 
Zhaokai Wang and Xue Yang are with Shanghai Jiao Tong University. \\ 
$^\dagger$ Corresponding author: Jifeng Dai (daijifeng@tsinghua.edu.cn).
}  }


% \markboth{Journal of \LaTeX\ Class Files,~Vol.~14, No.~8, August~2015}%
% {Shell \MakeLowercase{\textit{et al.}}: Bare Demo of IEEEtran.cls for Computer Society Journals}

\markboth{Mono-InternVL-1.5}%
{}




\IEEEtitleabstractindextext{%
\begin{abstract}
This paper focuses on monolithic Multimodal Large Language Models (MLLMs), which integrate visual encoding and language decoding into a single model.  Existing  structures and pre-training strategies for monolithic MLLMs often suffer from unstable optimization and catastrophic forgetting. To address these challenges, our key idea is to embed a new visual parameter space into a pre-trained LLM, enabling stable learning of visual knowledge from noisy data via delta tuning.  Based on this principle, we first introduce Mono-InternVL, an advanced monolithic MLLM that incorporates a set of visual experts through a multimodal mixture-of-experts architecture.  In addition, we design an innovative Endogenous Visual Pre-training (EViP) for Mono-InternVL  to maximize its visual capabilities via progressive learning.  Mono-InternVL  achieves competitive performance against existing MLLMs but also leads to relatively expensive  data cost.   Therefore, we further present Mono-InternVL-1.5, a cheaper and stronger monolithic MLLM equipped with an improved EViP (EViP++).  EViP++ introduces additional visual attention experts to  Mono-InternVL-1.5 and re-organizes the pre-training process in an efficient manner.  During inference, Mono-InternVL-1.5  includes a fused CUDA kernel to speed up its MoE operations.
With these designs, Mono-InternVL-1.5 significantly reduces training and inference costs, while still maintaining competitive performance with Mono-InternVL.  To evaluate our approach, we conduct extensive experiments across 15 benchmarks. Results demonstrate that Mono-InternVL  outperforms existing monolithic MLLMs on 12 out of 15 benchmarks, \emph{e.g.,} +114-point improvement over Emu3 on OCRBench. Compared to its modular counterpart, \emph{i.e.,} InternVL-1.5, Mono-InternVL-1.5 achieves similar multimodal performance while reducing first-token latency by up to 69\%. Code and models are released at {\small\url{https://github.com/OpenGVLab/Mono-InternVL}}.

\end{abstract}

}

\maketitle

\IEEEdisplaynontitleabstractindextext

% \fi
\IEEEpeerreviewmaketitle

\section{Introduction}\label{sec:introduction}
\label{sec:intro}

\begin{figure*}[t]
    \centering
    \includegraphics[width=\linewidth]{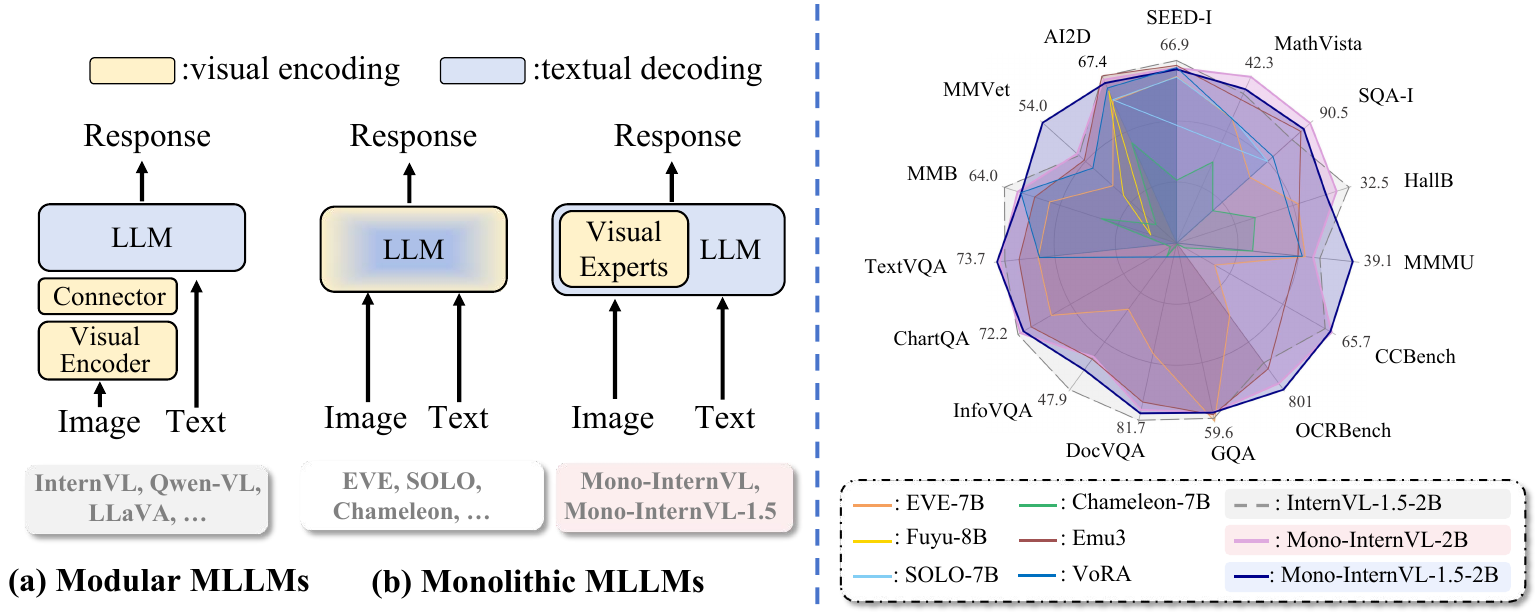}
    {
    \vspace{-0.5em}
    \caption{\textbf{Comparison of  Mono-InternVL, Mono-InternVL-1.5  and  existing MLLMs.} 
     Compared with modular MLLMs, Mono-InternVL and Mono-InternVL-1.5 embed visual experts into the pre-trained LLM and integrates visual encoding and language decoding into a single LLM. Through endogenous visual pre-training (EViP), Mono-InternVL significantly pushes the performance boundaries of monolithic MLLMs. With EViP++, Mono-InternVL-1.5 not only significantly reduces data costs, but also maintains the competitive performance of downstream tasks. 
    \label{fig:fig1}
    }
    % \vspace{-0.5em}
    }
\end{figure*}

Recent years have witnessed the  significant achievement  of Multimodal Large Language Models (MLLMs)~\cite{VLM:GPT-4, TransF:Qwen, cai2024internlm2} in various vision-language tasks. As illustrated in Fig.~\ref{fig:fig1}(a), most existing Multimodal Large Language Models (MLLMs) adopt a modular architecture, where visual encoding and language decoding are handled separately. This approach is typically realized by combining a pre-trained visual encoder~\cite{VLP:CLIP}  with an LLM~\cite{VLM:LLaVA, VLM:InternVL-1.5, VLP:BLIPv2}. In contrast, monolithic MLLMs~\cite{VLM:Fuyu-8b, diao2024EVE, solo} have become another popular research trend in the community, as shown in Fig.~\ref{fig:fig1}(b), which integrate visual perception and multimodal understanding within a unified LLM framework. Compared to modular MLLMs, monolithic MLLMs often  exhibit better potential in terms of design simplicity and deployment efficiency~\cite{diao2024EVE, solo}.

\begin{table}[t]
\caption{\textbf{Overall comparison of Mono-InternVL and Mono-InternVL-1.5.} Mono-InternVL-1.5 greatly improves the training and inference efficiency while maintaining competitive downstream performance. }
\begin{tabular}{lcccc}
\toprule
\multirow{2}{*}{Method} & Training & Inference & VQA    & MLLM  \\
                        & Data     & Speed     & Bench  & Bench \\ \midrule
Mono-InternVL~\cite{mono_internvl}           & 1.1B     & 61 tokens/s    & 70.1   & 53.7  \\
Mono-InternVL-1.5       & 0.5B     & 77 tokens/s    & 70.4   &   55.6     \\
$\Delta$                   & \textbf{-58\%}    & \textbf{+26\% }    & \textbf{+0.3\%} &    \textbf{+1.9\%}    \\ \bottomrule
\end{tabular}
\label{tab:summary}
\end{table}

Despite these advancements, training a monolithic MLLM that achieves competitive performance still remains a significant challenge. 
Among them,  \textit{native pre-training}~\cite{team2024chameleon}  pre-trains a monolithic MLLM from scratch using a combination of text-only and multimodal data.   However, this method demands extremely high computational resources and is prone to optimization instability~\cite{team2024chameleon}.  Another promising solution is to  extend the pre-trained LLM  to multimodality via additional visual pre-training, namely \textit{continuous pre-training}~\cite{diao2024EVE}. Such approaches typically require much cheaper training costs but easily incurs the catastrophic forgetting issue~\cite{zhai2023investigating}, thereby undermining the pre-trained language knowledge.

In this paper, we  aim to address the forgetting issue of continuous pre-training from the perspective of  \textit{delta tuning}~\cite{ding2022delta}.  Specifically, delta tuning  fine-tunes a set of newly added parameters in the model while keeping the rest frozen, thereby preserving the original knowledge.  However, existing methods adopt a shared architecture for joint vision and language modeling,  where optimizations for vision can negatively impact language capabilities. 
Therefore, it is a natural thought to introduce an independent visual parameter set into the pre-trained LLM, thus retaining the language knowledge by freezing the entire LLM  while facilitating visual learning.
This principle is also aligned with previous endeavors in modular MLLMs, \emph{e.g.,} QwenVL~\cite{bai2023qwenvl} and InternVL~\cite{VLM:InternVL-1.5},  where the visual parameters are placed outside the LLM.

\begin{figure*}[t!]
    \centering
    \includegraphics[width=\linewidth]{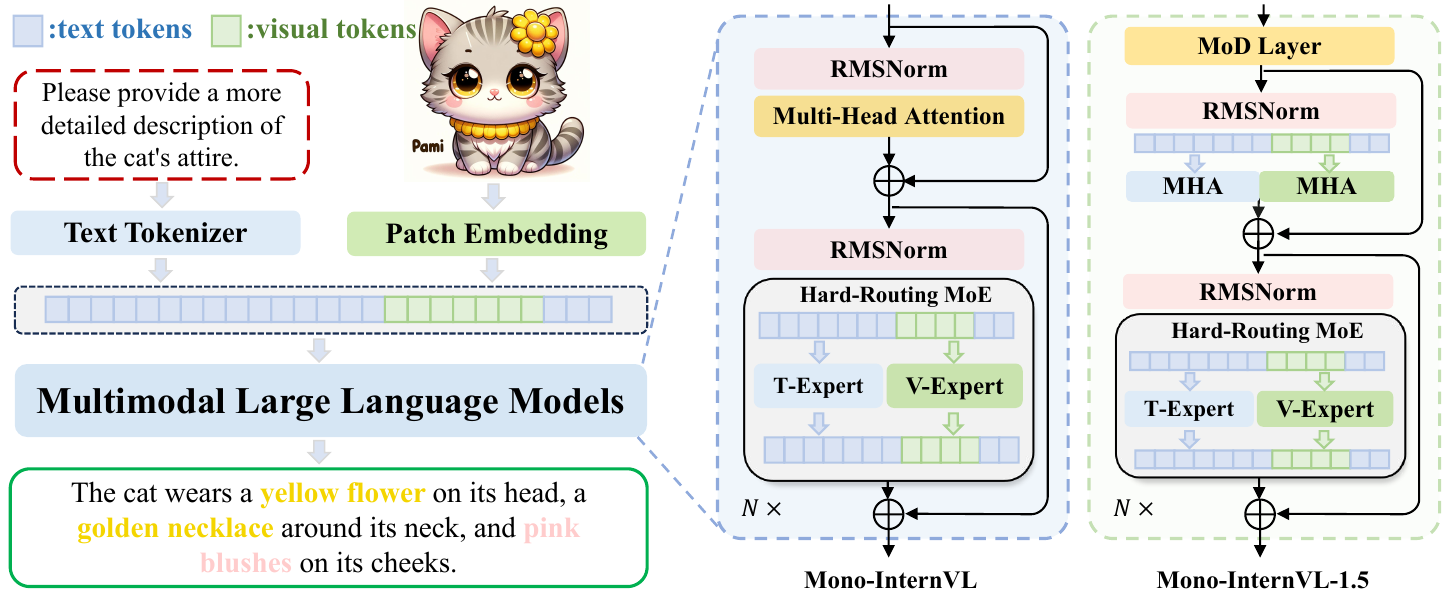}
    {
    \caption{\textbf{Monolithic architecture of Mono-InternVL and Mono-InternVL-1.5.}    Mono-InternVL is designed as a multimodal MoE structure, where  visual  and textual tokens are processed by the corresponding experts.  Mono-InternVL-1.5 further integrates the attention experts and the MoE CUDA kernel to   facilitate the visual pre-training while retaining the model efficiency.  }
    \label{fig:framework}
    }
\vspace{-8pt}
\end{figure*}

Based on the above principle, we propose a novel monolithic MLLM, namely Mono-InternVL. As shown in Fig.~\ref{fig:framework}, the visual parameters in Mono-InternVL are instantiated as a set of expert networks  via the mixture-of-experts (MoEs) mechanism.  Based on this architecture, we present an innovative \textit{Endogenous Visual Pre-training} (EViP) method to optimize the visual parameters. Specifically, EViP is formulated as a progressive learning process of three stages: 1) concept learning to grasp basic visual concepts, 2) semantic learning to capture high-level semantics, \emph{e.g.,} world knowledge, and 3) alignment learning to align knowledge with downstream tasks.   Benefiting from the architecture and the pre-training strategy, the visual scalability of Mono-InternVL is fully unleashed,  where the downstream performance consistently improves as the scale of the pre-training data increases.

Nevertheless, Mono-InternVL still requires expensive expenditures for its pre-training,  \emph{e.g.,} billions of image-text pairs, and its deployment is still unfriendly due to the modality-specific MoEs.  To overcome above limitations, we further  present Mono-InternVL-1.5, a cheaper and faster monolithic MLLM equipped with an improved Endogenous Visual Pre-training (EViP++).  Compared to EViP, the core idea of EViP++ is to maximize the learning ability of the model while minimizing redundancy of the data.
In particular, EViP++ firstly enlarges the visual parameter space and learning capability  by embedding visual attention experts into Mono-InternVL-1.5.  Then, EViP++ reorganizes the training data according to the principle of ``less is more''~\cite{zhou2023lima}, \textit{i.e.,} small in quantity but high in quality.   To further facilitate the efficiency, we  introduce a fused CUDA kernel to speed up the computation of the multimodal  mixture-of-experts mechanism.  As shown in Tab.~\ref{tab:summary}, the training data and inference cost of Mono-InternVL-1.5 can be significantly reduced, while the performance is still improved.

To validate our method, we develop Mono-InternVL and Mono-InternVL-1.5 using the pre-trained LLM InternLM2-1.8B~\cite{cai2024internlm2}, and conduct extensive experiments on 15 multimodal benchmarks.  Experimental results  demonstrate the significant performance improvements of  Mono-InternVL and Mono-InternVL-1.5 against previous monolithic MLLMs. For instance, Mono-InternVL-1.5 with 1.8 billion activated parameters  can obviously outperform existing monolithic MLLMs with 8 billion parameters, \emph{e.g.,} +2.8\%  over Emu3~\cite{emu3} on average.   Compared to the modular baseline, \emph{i.e.,} InternVL-1.5~\cite{VLM:InternVL-1.5}, Mono-InternVL-1.5 shows comparable performance on 15 multimodal benchmarks while  reducing  first token latency by  69.3\%.   In conclusion, our contributions can be summarized in five aspects:
\begin{itemize}[leftmargin=*]
    \item We present Mono-InternVL,  a novel monolithic MLLM that seamlessly integrates a set of  visual experts via  a multimodal mixture-of-experts architecture.  This architecture effectively extends the pre-trained LLM to a monolithic MLLM  while retaining the pre-trained knowledge.
    \item We propose a novel visual pre-training approach for Mono-InternVL called endogenous visual pre-training (EViP).  EViP adopts a progressive learning strategy to encourage visual experts to continuously grasp visual knowledge from noisy data to high-quality data.
    \item We introduce visual attention experts and an improved EViP (EViP++) to  boost the data efficiency during pre-training. Based on these strategies,  we present Mono-InternVL-1.5, a cheaper and faster monolithic MLLM that achieves stronger  performance than Mono-InternVL using only 42\%  data. 

    \item  We propose an innovative fused cuda kernel for the multimodal MoE in Mono-InternVL and Mono-InternVL-1.5, which greatly speeds up the model inference by up to 26\%. 
    
    \item Extensive experiments on 15 multimodal benchmarks  demonstrate that  our  monolithic MLLMs can reach the  comparable  performance and superior  efficiency  to  leading modular MLLMs, opening new avenues for designing future MLLMs.   
\end{itemize}

This paper is built upon our work published in CVPR 2025~\cite{mono_internvl}. 
Compared to the original version, we have made substantial extensions in five aspects in terms of model designs and experiments.
\textbf{1)} We present  Mono-InternVL-1.5, a cheaper  and faster monolithic MLLM than the original Mono-InternVL.  Mono-InternVL-1.5 demonstrates stronger downstream performance than Mono-InternVL on multiple MLLM benchmarks. \textbf{2)} In  Mono-InternVL-1.5, we introduce visual attention experts and an improved endogenous visual pre-training (EViP++) to significantly improve the data efficiency while retaining powerful performance.  Fig.~\ref{tab:datasets_all} illustrates the scaling property and advantages of EViP++.
\textbf{3)} We propose a novel CUDA kernel for multimodal mixture-of-experts, which can obviously speed up inference.  Our comparison in Tab.~\ref{tab:op-latency} confirms its advantages against the default Pytorch implementation.
\textbf{4)} In Tab.~\ref{ablation_attention} - \ref{ablation:shared}, we conduct more ablations and qualitative analysis to further compare  the impact of different designs in Mono-InternVL. \textbf{5)} In Tab.~\ref{tab:vqa_benchmark} - \ref{ablation_mmoe}, \ref{NLP_TABLE}, \ref{tab:speed} and Fig.~\ref{fig:data_curve_full} and \ref{vis:attention}, we conduct extensive experiments and visualizations  to validate Mono-InternVL-1.5 in terms of effectiveness and efficiency.  

\section{Related Work} 

\textbf{Modular multimodal large language models.~}
Recent advancements in large language models (LLMs) have driven the fusion of vision and language modalities, resulting in the development of multimodal large language models (MLLMs)~\cite{VLM:GPT-4v, VLM:Gemini,VLM:LLaVA-1.5,llava-hr,VLM:InternVL-1.5,piip_v2,emu3,SynerGen-VL}. Both commercial models like GPT-4o~\cite{VLM:GPT-4v} and Gemini series~\cite{VLM:Gemini} and open-source ones like BLIP series~\cite{VLP:BLIP,VLP:BLIPv2,VLM:InstructBLIP}, LLaVA series~\cite{VLM:LLaVA,VLM:LLaVA-1.5,VLM:LLaVA-1.6}, Qwen-VL~\cite{bai2023qwenvl, Qwen2vl,bai2025qwen2} and InternVL~\cite{VLM:InternVL,VLM:InternVL-1.5, InternVL-2.5, MPO, Mini-InternVL} have been actively working on this fusion. They often link LLMs~\cite{TransF:LLaMA,TransF:LLaMA2,cai2024internlm2, TransF:Qwen} with large vision models (LVMs)~\cite{VLP:CLIP,TransF:ViT,VLM:InternVL} through intermediate layers. 
Leveraging the advantages of extensively pre-trained visual encoders and state-of-the-art language models, these modular structures exhibit impressive performance across a broad range of multimodal tasks. Recent open-source frameworks, \emph{e.g.} InternVL 2.5~\cite{InternVL-2.5} and Qwen2.5-VL~\cite{bai2025qwen2}, demonstrate the efficacy of modular designs. Through large-scale multimodal pre-training and advanced visual-language alignment techniques, they achieve outcomes on par with leading commercial models.
However, as noted in~\cite{diao2024EVE}, such encoder-based vision-language models confront several issues. These include restrictions in visual processing due to pre-trained encoders, inefficiencies in deployment, and difficulties in balancing the capabilities of LLMs and LVMs.

\textbf{Monolithic multimodal large language models.~}
The problems linked to modular MLLMs have directed research towards encoder-free architectures, also referred to as monolithic MLLMs, which can be divided into two types. The first type centers on generating continuous visual tokens via lightweight structures prior to inputting them into MLLMs. For instance, Fuyu-8B~\cite{VLM:Fuyu-8b} processes images directly using a simple linear projection, adeptly handling high-resolution input images without requiring a specialized visual encoder. EVE-7B~\cite{diao2024EVE} emphasizes vision-language pre-alignment from an LLM-focused perspective and improves image recognition via visual distillation. SOLO~\cite{solo} puts forward an open-source training approach to facilitate the advancement of monolithic MLLMs.   In comparison, the second type introduces models based on VQ tokenizers to generate discrete visual tokens for image creation. Representative works include Chameleon~\cite{team2024chameleon}, Show-o~\cite{xie2024show}, Transfusion~\cite{zhou2024transfusion}, and Emu3~\cite{emu3}. These models convert images into discrete tokens, which simplifying the processing of visual information and enhancing generative capabilities. Monolithic MLLMs offer benefits such as not depending on pre-trained visual encoders, simplicity in design, and efficiency in deployment. Nonetheless, training a high-performance monolithic MLLM is still a significant challenge.

\textbf{Multimodal mixture-of-experts.~}
VLMo~\cite{bao2022vlmo} and BEiT-3~\cite{VLP:BEiTv3} use a set of modality experts to replace the feed-forward network in the Transformer. They effectively capture modality-specific information by switching to different modality experts and employ shared self-attention across modalities to align visual and linguistic information.
Based on the above works, VL-MoE~\cite{shen2023scaling} introduces mixture-of-experts (MoE)~\cite{yuksel2012twenty} to enhance efficiency of training and deployment.
MoMa~\cite{lin2024moma} also utilizes multimodal mixture-of-experts for pre-training MLLMs~\cite{team2024chameleon} and collaborates with sparse components, such as MoE and mixture-of-depths (MoD)~\cite{raposo2024mixture}, to boost the efficiency of pre-training from scratch with trillions of mixed-modal tokens. ARIA~\cite{li2024aria} further makes use of fine-grained multimodal MoEs to aid in understanding inputs from various data distributions, showcasing the potential of MoE architectures in constructing powerful MLLMs. Drawing inspiration from the above literature, we propose integrating multimodal mixture-of-experts (specifically, a visual expert and a language expert) into both multi-head attentions and feed-forward networks for pre-training monolithic MLLMs. We also introduce novel progressive learning strategies, namely endogenous visual pre-training (EViP and EViP++), to address the unique challenges of training monolithic MLLMs.

\section{Mono-InternVL} 

\subsection{Monolithic Architecture}

As illustrated in Fig.~\ref{fig:framework}, we first present the architecture of Mono-InternVL, which comprises tokenizers and a multimodal mixture-of-experts structure.

\textbf{Visual and textual embeddings.} In contrast to modular MLLMs,  Mono-InternVL directly convert images to input visual sequences with a lightweight patch embedding module.  Specifically, given the input image $I\in \mathbb{R}^{H\times W \times 3}$, the input visual embedding $x_v \in \mathbb{R}^{(h \times w) \times d}$ is obtained by
\begin{equation}
\begin{aligned}
x_v=\text{MLP}(\text{PatchEmbed}(I)+ \text{PE}).
\end{aligned}
\end{equation}
Here, $\text{PatchEmbed}(\cdot)$  denotes a patch embedding layer with a stride of 28, \emph{i.e.} each visual token correspond to a $28\times 28$ image patch.   $\text{PE} \in \mathbb{R}^{(h \times w) \times d}$ is the learnable positional embedding, similar to that in InternVL-1.5~\cite{VLM:InternVL-1.5}. We also add an additional thumbnail to provide global visual information into the model. 
Subsequently, an MLP layer $\text{MLP}(\cdot)$ is employed to project visual patches into the $d$-dimensional embedding space of the LLM.  This simple visual tokenizer enables Mono-InternVL to process images of arbitrary resolution with up to 8 millions of pixels, equivalent to $10,240$ image patches, covering most high-resolution scenarios. 

In Mono-InternVL, the textual tokenizer remains unchanged from the original one in the LLM. Given  the input text $T \in \mathbb{Z}^n$,   we obtain textual embedding $x_t \in \mathbb{R}^{n\times d}$ by  
\begin{equation}
\begin{aligned}
x_t=\text{Tokenizer}(T).
\end{aligned}
\end{equation}
Afterward,  the multimodal embedding is constructed by concatenating visual and textual embeddings, denoted as $x_m \in \mathbb{R}^{n' \times d}$.

\textbf{Multimodal mixture-of-experts structure.}
The core idea of Mono-InternVL is to embed visual experts into a pre-trained LLM. This allows Mono-InternVL not only to facilitate visual pre-training by leveraging the pre-trained LLM knowledge but also to significantly alleviate the catastrophic forgetting problem during pre-training.
Specifically, given  the multimodal input $x_m \in \mathbb{R}^{n' \times d }$,  a decoder-only LLM  with a set of visual experts is utilized to generate the  textual tokens step by step, which can be formulated by 
\begin{equation}
    \begin{aligned}
        p_s= \mathcal{F_\text{llm}}(y_{s}| x_m, y_{0:s-1};\theta,\theta_v).
    \end{aligned} 
    \label{eq_arch}
\end{equation}
Here, $y \in \mathbb{R}^{S}$  and $S$ denote the word length   and its length, respectively.  $p_s\in \mathbb{R}^{m}$ is the  next-token probability and $m$ is the size of the word vocabulary. $\mathcal{F}_\text{llm}$  and $\theta$ denote the LLM  and its  pre-trained parameters, respectively. $\theta_v$ refers to the parameters of the patch embedding layer and  visual experts. 

As shown in Fig.~\ref{fig:framework},  $\mathcal{F}_\text{llm}$ is designed as a multimodal mixture-of-experts structure.  Specifically, we adopt a static routing strategy that assigns visual and textual experts to their corresponding tokens. Therefore, the \textit{l}-th LLM layer  can be defined by
\begin{equation}
    \begin{aligned}
        x_m^{l'}&=x_m^{l-1}+\text{MHA}(\text{RMSNorm}(x_m^{l-1})),\\
        x_m^{l}&=x_m^{l'}+\text{MMoE}(\text{RMSNorm}(x_m^{l'})).
    \end{aligned}
    \label{mmoe_layer}
\end{equation}
Here, $\text{MHA}(\cdot)$ and $\text{RMSNorm}(\cdot)$ denote the multi-head attention~\cite{TransF:Transformer} and the layer normalization~\cite{zhang2019root}, respectively.  $\text{MMoE}(\cdot)$ is the proposed multimodal mixture-of-experts, formulated as
\begin{equation}
\begin{aligned}
    \text{MMoE}(x)=
\begin{cases} 
\text{FFN}_v(x) \quad \text{if } x \in x_v, \\
\text{FFN}_t(x) \quad \text{if } x \in x_t.
\end{cases}
\end{aligned}
\label{mmoe_module}
\end{equation}
 Here, $x \in \mathbb{R}^d$ is the element of $x_m$. $\text{FFN}_v$ and $\text{FFN}_t$ denote the visual and textual experts, respectively.   In practice, $\text{FFN}_v$ is initialized from the  $\text{FFN}_t$ to utilize the pre-trained knowledge.  

As defined in Eq.~\ref{mmoe_layer} and \ref{mmoe_module}, the MMoE structure has two distinct advantages over the existing monolithic MLLMs. Firstly, the visual learning of Mono-InternVL can largely benefit from the pre-trained language knowledge, while the language ability can still be preserved by freezing $\text{FFN}_t$. Secondly, the MMoE structure significantly improves the model's capacity for vision-and-language modeling, and the additional inference cost is almost negligible due to the MoE mechanism.

\begin{figure*}[t!]
    \centering
    \includegraphics[width=\linewidth]{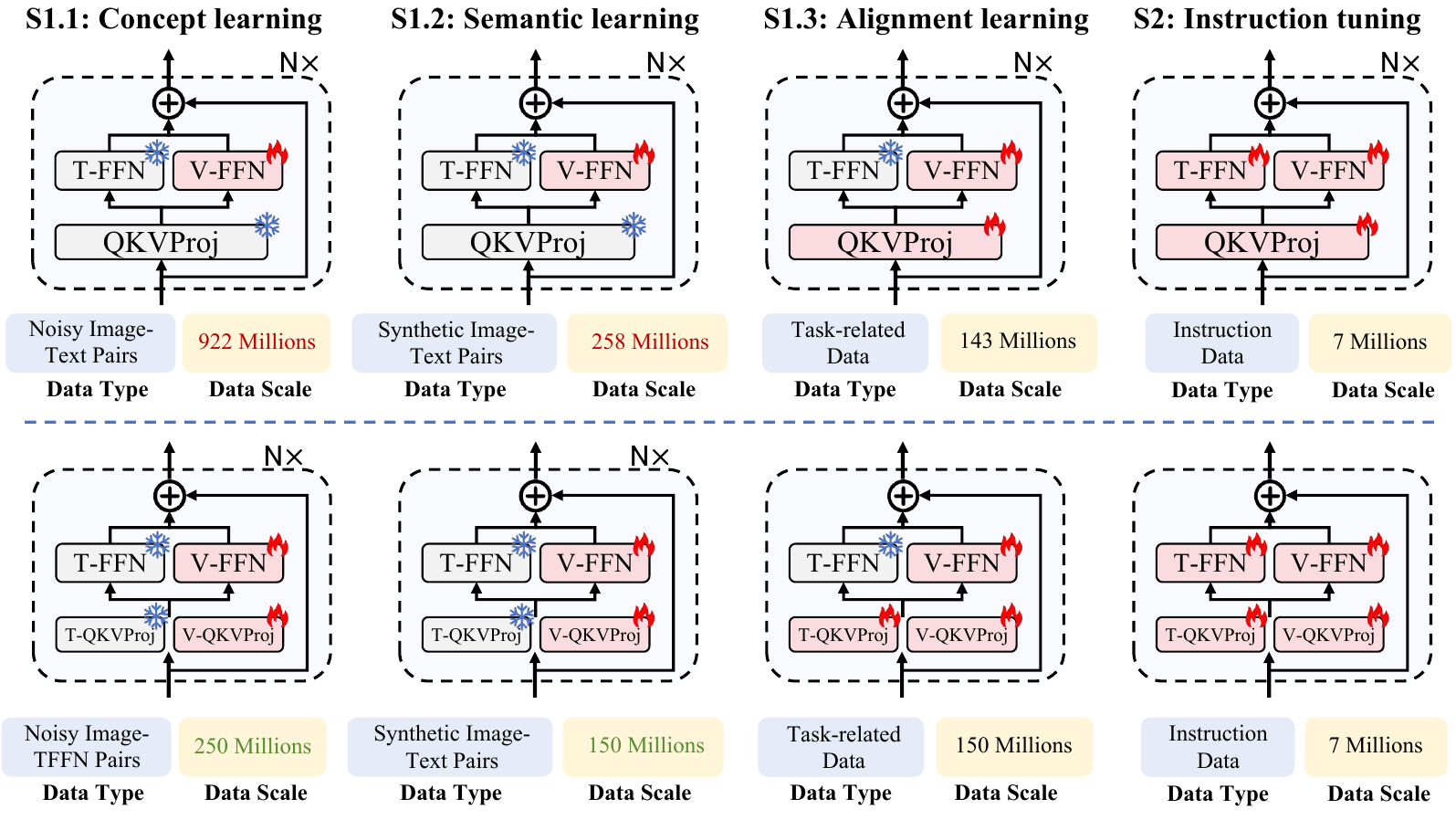}
    {
    % \captionsetup{belowskip=-0.5em}
    \vspace{-1em}
    \caption{\textbf{The training recipe of Mono-InternVL (top) and Mono-InternVL-1.5 (bottom).}  In the first stage, Mono-InternVL is progressively pre-trained on massive data via three sub-stages (S1.1, S1.2, S1.3), where most parameters of LLM are frozen to preserve the pre-trained knowledge. In the second stage (S2), the entire model is optimized to accommodate various instructions.  Compared to Mono-InternVL, Mono-InternVL-1.5  integrates  visual attention experts and reduces up to 58\% training data.  }
    \label{fig:fig2}
    }
% \vspace{-8pt}
\end{figure*}

\subsection{Endogenous Visual Pre-training}

The aim of Endogenous Visual Pre-training (EViP) is to maximize the benefits of Mono-InternVL from visual experts through pre-training on a large amount of noisy and synthetic data.  Unlike existing methods~\cite{diao2024EVE,team2024chameleon}, we formulate EViP from the perspective of delta tuning~\cite{ding2022delta}, where most LLM parameters are frozen to preserve the pre-trained knowledge.  Therefore, the objective of EViP can be defined as
\begin{equation}
    \begin{aligned}
        \arg \min_{\Delta \theta} \mathcal{L}(\mathcal{F}_{\text{llm}}(x_m; \theta, \theta_v), \hat{y}),
    \end{aligned}
    \label{delta_tuning}
\end{equation}
where $\mathcal{L}(\cdot)$ and $\hat{y}$ denote the auto-regressive loss and the ground truth, respectively. As shown in Fig.~\ref{fig:fig2}, $\Delta \theta$ represents the parameters of patch embedding and visual experts in the concept and semantic learning stages, \emph{i.e.,} $\theta_v$, while in the alignment learning stage, $\Delta \theta$ also includes the parameters of multi-head attentions. 
Based on Eq.~\ref{delta_tuning}, EViP is designed as a progressive learning process.  As illustrated in Fig.~\ref{fig:fig2} and Tab.~\ref{tab:datasets_all}, EViP consists of  three sub-stages, namely concept learning (S1.1), semantic learning (S1.2) and alignment learning (S1.3).  We use carefully partitioned data for each stage to achieve coarse-to-fine visual learning. 

\textbf{Concept learning.}  Concept learning aims to enable the model to learn basic visual concepts, such as object categories or basic shapes.  Therefore, we first pre-train Mono-InternVL with  about 922 million noisy data sampled from Laion-2B~\cite{Datasets:Laion-5b} and Coyo-700M~\cite{kakaobrain2022coyo-700m}.   In this sub-stage, Mono-InternVL uses a simple prompt for generative learning, \emph{i.e.,} ``provide a one-sentence caption for the image''. We limit the maximum   number of image patches of the  visual tokenizer to 1,280  for training efficiency. To preserve the language capabilities while enabling visual specialization, the entire LLM is frozen during concept learning, and only  the patch embedding and visual experts  are optimized. 

\textbf{Semantic learning.} After concept learning, Mono-InternVL can understand  basic concepts in the image, but it is still challenging to organize this information to generate reasonable descriptions.  To achieve a higher-level visual understanding,  we utilize the pre-trained InternVL2-8B~\cite{VLM:InternVL-1.5} to generate short captions for 258 million images.  Compared to the noisy captions in concept learning,  synthetic captions provide complex visual knowledge like relationship and world knowledge, \textit{etc.}, while containing less noisy information unrelated to the image, such as the shooting time or the photographer. In this sub-stage, we adopt the same optimization strategy as in concept learning, except that the maximum number of image patches is increased to 1,792.

\begin{table*}[t]
\scriptsize
\caption{\textbf{Summary of datasets used in the endogenous visual pre-training and instruction finetuning.}  In S1.2,  caption for each image is synthetically produced by the pre-trained InternVL2-8B~\cite{VLM:InternVL-1.5}.
}
\centering
\renewcommand{\arraystretch}{1.2}
\resizebox{\linewidth}{!}{
\begin{tabular}{cl}
\toprule
 \textbf{Stage} & \textbf{Datasets} \\ % \textbf{\#Samples} & 
 \midrule
S1.1  & Laion-EN (en)~\cite{Datasets:Laion-5b}, COYO (en)~\cite{kakaobrain2022coyo-700m}\\ % &  922M
  \midrule
  S1.2 & Laion-EN (en)~\cite{Datasets:Laion-5b}, COYO (en)~\cite{kakaobrain2022coyo-700m}, SAM (en)~\cite{TransF:SAM}        \\ %  &  258M 
  \midrule
  & \textbf{Captioning:} Laion-EN (en)~\cite{Datasets:Laion-5b}, Laion-ZH (zh)~\cite{Datasets:Laion-5b}, COYO (zh)~\cite{kakaobrain2022coyo-700m}, GRIT (zh)~\cite{peng2023kosmos2}, COCO (en)~\cite{chen2015cococaption}, TextCaps (en)~\cite{Datasets:TextCaps} \\
  & \textbf{Detection:} Objects365 (en\&zh)~\cite{shao2019objects365}, GRIT (en\&zh)~\cite{peng2023kosmos2}, All-Seeing (en\&zh)~\cite{wang2023allseeing} \\
  & \textbf{OCR (large):} Wukong-OCR (zh)~\cite{gu2022wukong}, LaionCOCO-OCR (en)~\cite{Datasets:LAION-COCO}, Common Crawl PDF (en\&zh) \\
  & \textbf{OCR (small):} MMC-Inst (en)~\cite{liu2023mmcinst}, LSVT (zh)~\cite{sun2019lsvt}, ST-VQA (en)~\cite{Datasets:STVQA}, RCTW-17 (zh)~\cite{shi2017rctw17}, ReCTs (zh)~\cite{zhang2019rects}, ArT (en\&zh)~\cite{chng2019art}, SynthDoG (en\&zh)~\cite{kim2022synthdog}, \\
  \multirow{-5}{*}{S1.3} & COCO-Text (en)~\cite{veit2016cocotext}, ChartQA (en)~\cite{Datasets:ChartQA}, CTW (zh)~\cite{yuan2019ctw}, DocVQA (en)~\cite{Datasets:DocVQA}, TextOCR (en)~\cite{singh2021textocr}, PlotQA (en)~\cite{methani2020plotqa}, InfoVQA (en)~\cite{mathew2022infographicvqa}    \\ % \multirow{-5}{*}{143M}  & 
  \midrule
  
    &   \textbf{Captioning:} TextCaps (en)~\cite{Datasets:TextCaps}, ShareGPT-4o (en\&zh)~\cite{VLM:InternVL-1.5}   \\
    & \textbf{General QA:} VQAv2 (en)~\cite{Datasets:VQAv2}, GQA (en)~\cite{Datasets:GQA}, OKVQA (en)~\cite{Datasets:Ok-vqa}, VSR (en)~\cite{liu2023vsr}, VisualDialog (en)~\cite{das2017visualdialog}     \\
    & \textbf{Science:} AI2D (en)~\cite{Datasets:AI2D}, ScienceQA (en)~\cite{Datasets:ScienceQA}, TQA (en)~\cite{kembhavi2017tqa}  \\
    & \textbf{Chart:} ChartQA (en)~\cite{Datasets:ChartQA}, MMC-Inst (en)~\cite{liu2023mmcinst}, DVQA (en)~\cite{kafle2018dvqa}, PlotQA (en)~\cite{methani2020plotqa}, LRV-Instruction (en)~\cite{liu2023lrv-instruction}   \\
    
    & \textbf{Mathematics:} GeoQA+ (en)~\cite{cao2022geoqa_plus}, TabMWP (en)~\cite{lu2022tablemwp}, MathQA (en)~\cite{yu2023mathqa}, CLEVR-Math/Super (en)~\cite{lindstrom2022clevrmath, li2023superclevr}, Geometry3K (en)~\cite{lu2021geometry3k} \\
     & \textbf{Knowledge:} KVQA (en)~\cite{shah2019kvqa}, A-OKVQA (en)~\cite{schwenk2022aokvqa}, ViQuAE (en)~\cite{lerner2022viquae}, Wikipedia (en\&zh)~\cite{he2023wanjuan} \\
     
    & \textbf{OCR:} OCRVQA (en)~\cite{Datasets:OCRVQA}, InfoVQA (en)~\cite{mathew2022infographicvqa}, TextVQA (en)~\cite{Datasets:TextVQA}, ArT (en\&zh)~\cite{chng2019art}, COCO-Text (en)~\cite{veit2016cocotext}, CTW (zh)~\cite{yuan2019ctw}, \\
    & LSVT (zh)~\cite{sun2019lsvt}, RCTW-17 (zh)~\cite{shi2017rctw17}, ReCTs (zh)~\cite{zhang2019rects}, SynthDoG (en\&zh)~\cite{kim2022synthdog}, ST-VQA (en)~\cite{Datasets:STVQA}  \\
    
    & \textbf{Document:} DocVQA (en)~\cite{Datasets:DocVQA}, Common Crawl PDF (en\&zh)                                                   \\
    & \textbf{Grounding:} RefCOCO/+/g (en)~\cite{Datasets:REFCOCO, Datasets:REFCOCOG}, Visual Genome (en)~\cite{Datasets:VisualGenome}                            \\
    
    & \textbf{Conversation:} LLaVA-150K (en\&zh)~\cite{VLM:LLaVA}, LVIS-Instruct4V (en)~\cite{wang2023lvisinstruct4v}, ALLaVA (en\&zh)~\cite{chen2024allava}, Laion-GPT4V (en)~\cite{laion_gpt4v_dataset}, ShareGPT (en\&zh)~\cite{zheng2023vicuna},  SVIT (en\&zh)~\cite{Datasets:SVIT}                            \\
    & \textbf{Text-only:} OpenHermes2.5 (en)~\cite{OpenHermes2_5}, Alpaca-GPT4 (en)~\cite{taori2023alpaca}, COIG-CQIA (zh)~\cite{bai2024coig}, ShareGPT (en\&zh)~\cite{zheng2023vicuna}                                   \\
    
     & \textbf{Video:} EgoTaskQA (en)~\cite{jia2022egotaskqa}, Mementos (en)~\cite{wang2024mementos}, STAR (en)~\cite{wu2021star_situated_reasoning}, NTU RGB+D (en)~\cite{shahroudy2016ntu}, VideoChat2IT (en\&zh)~\cite{li2023videochat}, LSMDC-QA (en)~\cite{lsmdc}, ShareGPT-4o (en\&zh)~\cite{VLM:InternVL-1.5} \\
    
    \multirow{-14}{*}{S2} & \textbf{Handwritten:} SROIE (en)~\cite{huang2019icdar2019}, FUNSD (en)~\cite{jaume2019}, POIE (en)~\cite{kuang2023visual} \\
    
  \bottomrule
\end{tabular}
\label{tab:datasets_all}
}
\end{table*}

\begin{figure}[t!]
    \centering
    \includegraphics[width=\linewidth]{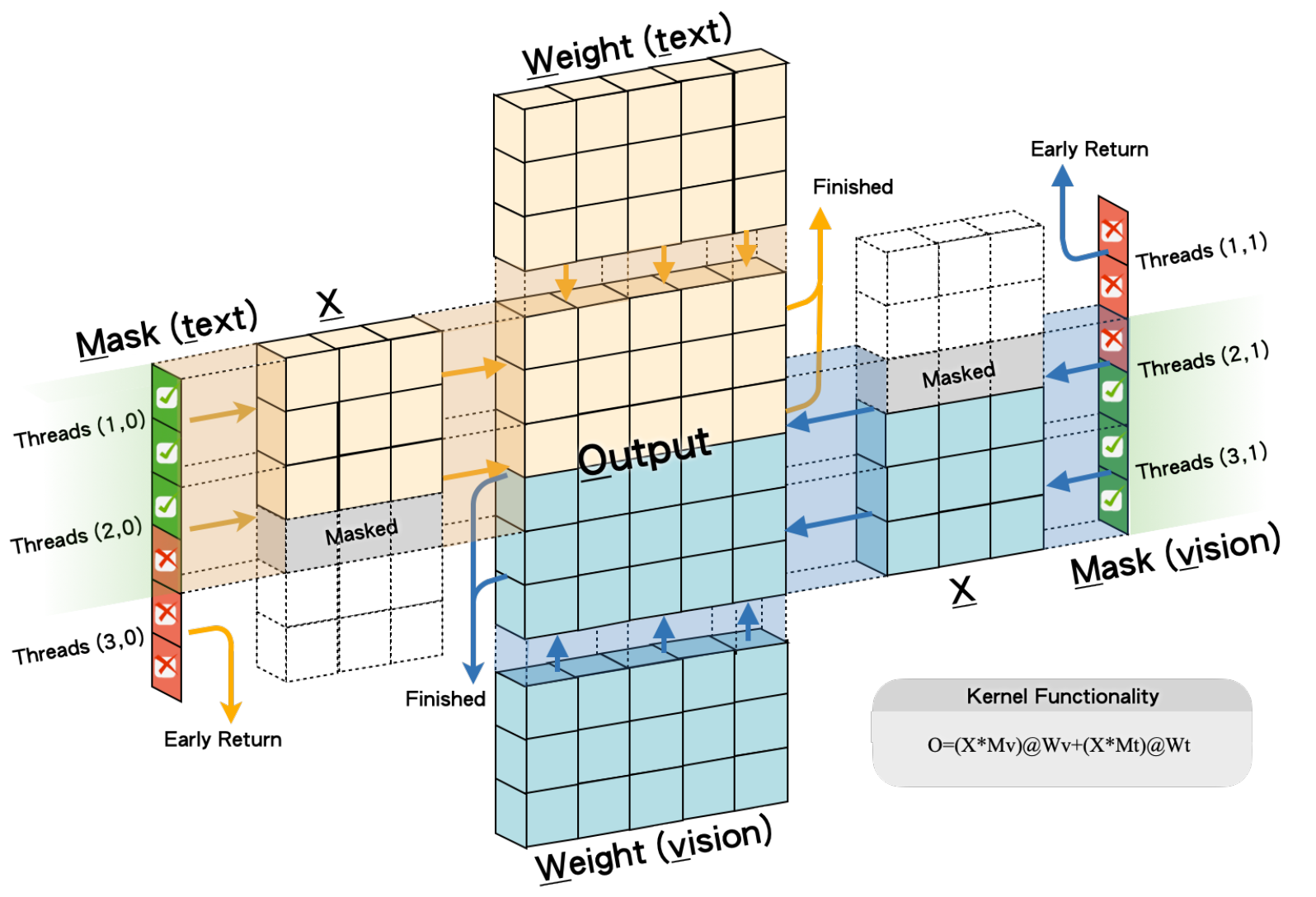}
    {
    % \captionsetup{belowskip=-0.5em}
    \vspace{-.2em}
    \caption{\textbf{Illustration of Mono-InternVL-1.5 fused kernel workflow.} The left thread blocks handle textual tokens while those on the right handle visual tokens. Although two thread blocks are assigned per data block, nearly half exit immediately upon entry, making the kernel effectively behave as a single-branch implementation.}
    \label{fig:kernel}
    }
\end{figure}

\textbf{Alignment learning.}   To improve the visual capability for downstream tasks, we adopt perform alignment learning on Mono-InternVL. Our alignment data are sampled from the pre-training data of InternVL-1.5~\cite{VLM:InternVL-1.5}, including 143 million  samples of image captioning, detection and optical character recognition (OCR), as shown in Tab.~\ref{tab:datasets_all}. Specifically, captioning, detection and OCR data account for about 53.9\%, 5.2\% and 40.9\% of the total total, respectively.  In this sub-stage, we use the task-specific prompts from InternVL-1.5 for the generative learning, and increase the maximum number of image patches to 3,328. Compared to previous sub-stages,  we additionally unfreeze the multi-head attention layers for better vision-language alignment.

\subsection{Instruction Tuning}
In this stage, we follow InternVL-1.5~\cite{VLM:InternVL-1.5} to perform supervised learning using around 7 million bilingual instructions, covering various tasks like visual question answering, multimodal dialogue, knowledge, mathematics,  \textit{etc}. 
In this stage, the entire models are unfreezed, and the maximum number of image patches is increased to 6,400 to handle high-resolution images. We list details of instruction data in Tab.~\ref{tab:datasets_all}.

\section{Mono-InternVL-1.5} 

\subsection{Improved Endogenous Visual Pre-training}
\textbf{Visual attention experts.}  In S1.1 and 1.2, the learning capability of Mono-InternVL is limited since its attention layers are frozen and initialized with textual knowledge. However, directly fine-tuning the attention parameters will lead to the catastrophic forgetting of textual knowledge.  Therefore, to further improve the learning capability of the model, Mono-InternVL-1.5 inserts additional visual experts into the multi-head attentions (MHA),  yielding a fully multimodal mixture-of-experts (MMoEs) architecture.    In particular, the $l$-th LLM layer can be rewritten as:
\begin{equation}
    \begin{aligned}
        x_m^{l'}&=x_m^{l-1}+\text{MMHA}(\text{RMSNorm}(x_m^{l-1})),\\
        x_m^{l}&=x_m^{l'}+\text{MMoE}(\text{RMSNorm}(x_m^{l'})).
    \end{aligned}
    \label{new_mmoe_layer}
\end{equation}
As shown in Fig.~\ref{fig:framework},  the calculation of $\text{MMHA}(\cdot)$ is similar to $\text{MHA}(\cdot)$, \emph{i.e.,} $\text{softmax}(a\cdot qk)v$, but when computing the \textit{query}, \textit{key} and \textit{value} in the attention, visual and textual tokens are assigned with different linear expert layers.  For example, given the input features $x \in \mathbb{R}^{l\times d}$, the computation of the \textit{query}  $q \in \mathbb{R}^{l\times d}$  can be defined as:
\begin{equation}
\begin{aligned}
    q=
\begin{cases} 
\text{Linear}_v(x) \quad \text{if } x \in x_v, \\
\text{Linear}_t(x) \quad \text{if } x \in x_t.
\end{cases}
\end{aligned}
\label{mmol_module}
\end{equation}
Similarly, we can obtain the \textit{key} $k \in \mathbb{R}^{l\times d}$ and the \textit{value} $v \in \mathbb{R}^{l\times d}$.  

With this architecture, we find that the training efficiency of Mono-InternVL-1.5 is significantly improved during the pre-training stage.  Furthermore, the inference efficiency can also be retained  through the MoE mechanism.

\begin{table*}[t!]
\scriptsize
\vspace{-2mm}
\caption{\textbf{Comparison with existing MLLMs on general MLLM benchmarks.}    ``\#A-Param'' denotes the number of activated parameters. Average scores are computed by normalizing each metric to a range between 0 and 100. $^\dagger$ InternVL-1.5-2B adopts the same LLM and high-quality training data with Mono-InternVL-2B, so we consider it as the modular counterpart. \textbf{Bold} indicates the highest performance among monolithic MLLMs.  
}
% \vspace{-2mm}
\centering
\setlength\tabcolsep{2pt}
\renewcommand{\arraystretch}{1.15}
\resizebox{1.\linewidth}{!}{
\begin{tabular}{lc|cccccccc|c}
\toprule
Model & \#A-Param & MMB & MMVet & MMMU  & MathVista & SEED-I & OCRBench & HallB & CCB & Avg \\
\midrule
\multicolumn{3}{l}{\emph{Modular MLLMs:}}   & & & & & & \\
MobileVLM-V2-3B~\cite{chu2024mobilevlm}           & 3.0B & 63.2 & $-$ & $-$ & $-$ & $-$ & $-$ & $-$ & $-$ & $-$ \\
Mini-Gemini-2B~\cite{VLM:MiniGemini}      & 3.5B & 59.8 & 31.1 & 31.7  & 29.4 & $-$ & $-$ & $-$ & $-$ & $-$ \\
MM1-3B-MoE-Chat~\cite{VLM:MM1} & 3.5B & 70.8 & 42.2 & 38.6  & 32.6 & 69.4 & $-$ & $-$ & $-$ & $-$ \\
DeepSeek-VL-1.3B~\cite{lu2024deepseekvl}   & 2.0B & 64.6 & 34.8 & 32.2 & 31.1 & 66.7 & 409 & 27.6 & 37.6 & 41.9 \\
PaliGemma-3B~\cite{beyer2024paligemma}  & 2.9B & 71.0 & 33.1 & 34.9 & 28.7  & 69.6 & 614 & 32.2 & 29.6 & 45.0 \\
MiniCPM-V-2~\cite{yao2024minicpm}   & 2.8B & 69.1 & 41.0 & 38.2  & 38.7  & 67.1 & 605 & 36.1 & 45.3 & 49.5 \\
$^\dagger$InternVL-1.5-2B~\cite{VLM:InternVL-1.5} & 2.2B & 70.9 & 39.3 & 34.6  & 41.1 & 69.8 & 654 & 37.5 & 63.5 & 52.7 \\
Qwen2VL-2B~\cite{Qwen2vl} & 2.1B & 74.9 & 49.5 & 41.1 & 43.0 &$-$& 809 & 41.7 &$-$&$-$\\
InternVL-2.5-2B~\cite{chen2024expanding} & 2.2B & 74.7 & 60.8 & 43.6  & 51.3 &$-$& 804 & 42.6 & $-$ & $-$\vspace{-0.6mm} \\
\midrule

\multicolumn{3}{l}{\emph{Monolithic MLLMs:}}   & & & & & & & \\
Fuyu-8B (HD)~\cite{VLM:Fuyu-8b} & 8B & 10.7 & 21.4 & $-$ & $-$ & $-$ & $-$ & $-$  & $-$ & $-$ \\
SOLO~\cite{solo}         & 7B & $-$ & $-$ & $-$   & 34.4 & 64.4 &  $-$ & $-$ & $-$ & $-$ \\
Chameleon-7B\footnotemark[1]~\cite{team2024chameleon} & 7B & 31.1 & 8.3 & 25.4  & 22.3 & 30.6 & 7 & 17.1 & 3.5 & 17.3 \\
EVE-7B~\cite{diao2024EVE}       & 7B & 49.5 & 25.6 & 32.3& 25.2 & 61.3 & 327 & 21.1 & 12.4 & 32.5 \\
EVE-7B (HD)~\cite{diao2024EVE}  & 7B & 52.3 & 25.7 & 32.6 & 34.2 & 64.6 & 398 & 26.4 & 16.3 & 36.5 \\
Emu3~\cite{emu3} & 8B & 58.5 & 37.2 & 31.6  & $-$ & 68.2  & 687& $-$ & $-$ & $-$\\ 
VoRA~\cite{vora}        & 7B & 64.2 & 33.7 & 32.2  &$-$& 67.5 &$-$&$-$&$-$&$-$\\
VoRA-AnyRes~\cite{vora} & 7B & 61.3 & 33.7 & 32.0  &$-$& \textbf{68.9 }&$-$&$-$&$-$&$-$\\
\rowcolor{gray!15}
Mono-InternVL-2B & 1.8B & \textbf{65.5} & 40.1 & 33.7  & \textbf{45.7} & 67.4 & 767 & \textbf{34.8} & \textbf{66.3} & {53.7} \\

\rowcolor{gray!15} Mono-InternVL-1.5-2B & 1.8B &  64.0   & \textbf{54.0} & \textbf{39.1} & 42.3 & 66.9 & \textbf{801} & 32.5 & 65.7  &  \textbf{55.6}\vspace{-0.6mm}  \\
\bottomrule
\end{tabular}
}
\label{tab:multimodal_benchmark}
\vspace{1mm}
\end{table*}

\textbf{Improved training strategies and data organization.}   In EViP, concept learning consumes almost billions of  data to learn basic visual concepts, leading to relatively expensive expenditure. However, through our empirical studies, the performance gain of concept learning grows slowly as the data scale increases.  On the one hand, only the visual experts are optimized during concept learning, which yields  suboptimal learning efficiency. On the other hand, most samples for concept learning are noisy and simple, so it is difficult for MLLM to quickly learn useful information from them.   Existing methods~\cite{zhou2023lima} also show that a small amount of high-quality data can achieve performance comparable to that of large-scale low-quality data.

Motivated by the above analysis,  we improve the training strategies and data organization of EViP from two aspects, as shown in Fig.~\ref{fig:fig2}.   Firstly, we  integrate visual attention experts into Mono-InternVL-1.5 and optimize their parameters during visual pre-training.  By doing so, the visual and textual modalities can be quickly aligned in multi-head attentions, thereby leading to better training efficiency.  Notably, similar to visual experts of Mono-InternVL,  the optimization of   visual attention experts will not affect the language capabilities.  Secondly, we re-organize the training data of EViP based on the principle in the existing literature~\cite{zhou2023lima}, \emph{i.e.,} less noisy data and more valuable data. Specifically, we reduce the pre-training data from 922 million and 258 million to 250 million and 150 million for S1.1 and S1.2, respectively.  Then, the data of S1.3 and instruction tuning  is slightly increased to compensate for model performance. 

With these strategies, the training data of Mono-InternVL-1.5 is reduced by about 58\%, while the downstream performance can still be improved.

\footnotetext[1]{Chameleon-7B frequently rejects to perform the task with a response of ``I can't help you with this", thus resulting in poor performance.}

\subsection{Speeding Up Mono-InternVL-1.5 with Fused CUDA Kernel}
As shown in Fig.~\ref{fig:framework}, Mono-InternVL-1.5 adopts a full multimodal MoE architecture, where visual and textual tokens are processed by two different experts, respectively. Unfortunately,  none of the mainstream frameworks or libraries  support the deployment of this modality-specific MoE.   In paractice, visual and textual tokens must be separated and processed sequentially, which limits GPU parallelism, especially during inference when the amount of data is relatively small.

To address this issue, we propose a fused CUDA kernel handles both branches jointly, thereby reducing latency and improving GPU utilization.
The core idea is based on the observation that, if the input sequence is partitioned into find-grained blocks, the likelihood that both token types co-occur within a single block becomes low. 
As illustrated in Fig.~\ref{fig:kernel}, we divide the sequence into smaller blocks and assign two thread blocks to each: one responsible for visual tokens and the other for textual tokens. Upon initialization, each thread block checks for the presence of relevant tokens, and if none are found, it exits immediately.

This design ensures that only a small portion of blocks require both thread blocks to be active, while the majority can be handled by a single thread block, thus closely approaching the efficiency of single-branch computation.

\begin{table*}[!t]
\scriptsize
\caption{\textbf{Comparison with existing MLLMs on  visual question answering benchmarks.}
}
% \vspace{-2mm}
\centering
\setlength\tabcolsep{5pt}
\renewcommand{\arraystretch}{1.15}
\resizebox{1.\linewidth}{!}{
\begin{tabular}{lc|ccccccc|cc}
\toprule
Model & \#A-Param & TextVQA & SQA-I & GQA & DocVQA & AI2D & ChartQA & InfoVQA & Avg \\
\midrule
\multicolumn{3}{l}{\emph{Modular MLLMs:}}  & & & & & \\
MobileVLM-V2-3B~\cite{chu2024mobilevlm}   & 3.0B & 57.5 & 70.0 & 66.1 & $-$ & $-$ & $-$ & $-$ & $-$ \\
Mini-Gemini-2B~\cite{VLM:MiniGemini}    & 3.5B & 56.2 & $-$ & $-$ & 34.2 & $-$ & $-$ & $-$ & $-$ \\
MM1-3B-MoE-Chat~\cite{VLM:MM1} & 3.5B & 72.9 & 76.1 & $-$ & $-$ & $-$ & $-$ & $-$ & $-$ \\
DeepSeek-VL-1.3B~\cite{lu2024deepseekvl}  & 2.0B & 57.8 & $-$ & $-$ & $-$ & 51.5 & $-$ & $-$ & $-$ \\
PaliGemma-3B~\cite{beyer2024paligemma}      & 2.9B & 68.1 & $-$ & $-$ & $-$ & 68.3 & $-$ & $-$ & $-$ \\
MiniCPM-V-2~\cite{yao2024minicpm}       & 2.8B & 74.1 & $-$ & $-$ & 71.9 & 62.9 & $-$ & $-$ & $-$ \\
$^\dagger$InternVL-1.5-2B~\cite{VLM:InternVL-1.5}   & 2.2B & 70.5 & 84.9 & 61.6 & 85.0 & 69.8 & 74.8 & 55.4 & 71.7 \\
Qwen2VL-2B~\cite{Qwen2vl} & 2.1B & 79.7 & $-$ & $-$ & 90.1 & 74.7 & 73.5 & 65.5 & $-$  \\
InternVL-2.5-2B~\cite{chen2024expanding} & 2.2B & 74.3 & $-$ & $-$ & 88.7 & 74.9 & 79.2 & 60.9 & $-$\vspace{-0.6mm} \\
\midrule
\multicolumn{3}{l}{\emph{Monolithic MLLMs:}}  & & & & & \\
Fuyu-8B (HD)~\cite{VLM:Fuyu-8b} & 8B & $-$ & $-$ & $-$ & $-$ & 64.5 & $-$ & $-$ & $-$ \\
SOLO~\cite{solo}         & 7B & $-$ & 73.3 & $-$ & $-$ & 61.4 & $-$ & $-$ & $-$ \\
Chameleon-7B\footnotemark[1]~\cite{team2024chameleon} & 7B & 4.8 & 47.2 & $-$ & 1.5 & 46.0 & 2.9 & 5.0 & 17.9 \\
EVE-7B~\cite{diao2024EVE}       & 7B & 51.9 & 63.0 & 60.8 & 22.0 & 48.5 & 19.5 & 20.0 & 40.8 \\
EVE-7B (HD)~\cite{diao2024EVE}  & 7B & 56.8 & 64.9 & \textbf{62.6} & 53.0 & 61.0 & 59.1 & 25.0 & 54.6 \\
Emu3~\cite{emu3}  & 8B & 64.7 & 89.2 & 60.3 & 76.3 & \textbf{70.0} & 68.6 & 43.8 & 67.6 \\
VoRA~\cite{vora}        & 7B & 56.3 & 75.9 & $-$ & $-$ &  65.6 & $-$ & $-$ & $-$ \\
VoRA-AnyRes~\cite{vora} & 7B & 58.7 & 72.0 & $-$ & $-$ & 61.1 & $-$ & $-$ & $-$ \\
\rowcolor{gray!15}
Mono-InternVL-2B & 1.8B & 72.6 & \textbf{93.6} & 59.5 & 80.0 & {68.6} & \textbf{73.7} & 43.0 & 70.1 \\
\rowcolor{gray!15} Mono-InternVL-1.5-2B & 1.8B & \textbf{73.7} & 90.5  & {59.6} &  \textbf{81.7}&  67.4 & 72.2&  \textbf{47.9} & \textbf{70.4}\vspace{-0.6mm} \\
\bottomrule
\end{tabular}
}
\label{tab:vqa_benchmark}
\end{table*}

\newpage

\section{Experiments}

\subsection{Evaluation Benchmarks}
\vspace{-1mm}
We evaluate Mono-InternVL and existing MLLMs on 15 comprehensive multimodal benchmarks and 4 natural language processing (NLP) benchmarks. 
Specifically, MLLM benchmarks include MMBench-EN \textit{test}~\cite{Datasets:MMBench}, MMVet~\cite{Datasets:MM-vet}, MMMU \textit{val}~\cite{Datasets:MMMU}, MathVista \textit{testmini}~\cite{Datasets:Mathvista}, SEED-Image~\cite{Datasets:Seed-bench}, OCRBench~\cite{liu2023ocrbench}, HallusionBench~\cite{Datasets:Hallusionbench}, and CCBench \textit{dev}~\cite{Datasets:MMBench}. Visual question answering benchmarks include TextVQA \textit{val}~\cite{Datasets:TextVQA}, SQA \textit{test}~\cite{Datasets:ScienceQA}, GQA \textit{test-dev}~\cite{Datasets:GQA}, DocVQA \textit{test}~\cite{Datasets:DocVQA}, AI2D \textit{test}~\cite{Datasets:AI2D}, ChartQA \textit{test}~\cite{Datasets:ChartQA}, and InfographicVQA \textit{test} \cite{mathew2022infographicvqa}. 
NLP benchmarks include MMLU~\cite{MMLU}, CMMLU~\cite{CMMLU}, AGIEval~\cite{AGIEVAL} and MATH~\cite{MATH}.
The evaluation metrics follow  existing methods~\cite{VLM:InternVL-1.5,diao2024EVE}.
Some results of Chameleon and EVE are evaluated with VLMEvalKit~\cite{duan2024vlmevalkit} or from the OpenCompass leaderboard~\cite{opencompass2023}.

\begin{table*}[t]
\footnotesize
\centering
\caption{\textbf{Zero-shot pre-training performance of Mono-InternVL and existing MLLMs.}  ``S1.2'' and ``S1.3'' denote pre-training stages of semantic learning and alignment learning, respectively. Images of COCO have been seen in Mono-InternVL-S1.3, so we mark its performance in gray.}
\renewcommand{\arraystretch}{1}
\resizebox{0.8\linewidth}{!}{
\begin{tabular}{lccc|ccccc}
\toprule
\multicolumn{1}{l}{Model} & \#A-Param   & Data  & Shots & COCO Caps  & Flickr30k & NoCaps  & VQAv2 \\ \midrule
Flamingo~\cite{alayrac2022flamingo}                    & 3B     &  \textgreater  2.1B    & 0     & 73.0  & $-$        & $-$           & 49.2  \\
MM1~\cite{mckinzie2024mm1}                         & 3.5B   & \textgreater  2.3B  & 0     & 73.5  & $-$  & 55.6     & 46.2  \\
Chameleon~\cite{team2024chameleon}                   & 34B   & \textgreater 1.4B   & 2     & 120.2 & 74.7      & $-$        & 66.0  \\
\rowcolor{gray!15} Mono-InternVL-S1.2             & 1.8B& 0.9B&  0     & 87.3   & 72.7          &  54.1      &   $-$           \\
\rowcolor{gray!15} Mono-InternVL-S1.3             & 1.8B &1.1B& 0     & \textcolor{gray}{135.6}      &  77.3         & 116.5       &        \textcolor{gray}{71.1}  \\ 
\rowcolor{gray!15} Mono-InternVL-1.5-S1.2    & 1.8B & 0.4B& 0 & 91.5 & 71.6 &53.5 & -        \\
\rowcolor{gray!15} Mono-InternVL-1.5-S1.3    & 1.8B& 0.5B
& 0 & \textcolor{gray}{133.0} & 76.7& 115.3 &\textcolor{gray}{70.7}\vspace{-0.6mm} \\ 
\bottomrule
\end{tabular}
}
\label{few_shot_result}
\end{table*}

\subsection{Implementation Details}
We build Mono-InternVL upon InternLM2-1.8B~\cite{cai2024internlm2} with newly added visual tokenizer and visual experts. For Mono-InternVL-1.5, visual attention experts are also added. The visual experts are initialized from  pre-trained MLPs in the original InternLM2-1.8B to utulize existing learned representations for improved visual feature extraction. The visual experts account for 1.2 billion parameters. Similarly, the visual attention experts are also initialized from the pre-trained attention weights in  InternLM2-1.8B.
For both Mono-InternVL and Mono-InternVL-1.5, we adopt a similar dynamic high-resolution strategy from InternVL-1.5~\cite{VLM:InternVL-1.5} to align an optimal resolution for input image, which is then patchified to visual tokens.  The other configurations are identical to InternLM2-1.8B. For Mono-InternVL, the endogenous visual pre-training  and instruction tuning take approximately 17 days   on 256 NVIDIA A100 GPUs.  For Mono-InternVL-1.5, the total training time is reduced to 9.5 days.

\begin{table*}[t]
\small
\centering
\caption{\textbf{Ablation of different strategies for visual pre-training.}  All models are pre-trained on 61 million image-text pairs from Laion-2B~\cite{Datasets:Laion-5b} and fine-tuned on instruction data from LLaVA-665k~\cite{VLM:LLaVA-1.5}.  ``Full'' and ``Delta'' denote full tuning and delta tuning, respectively.  ``T-Param'' refers to trainable parameters.} 
\renewcommand{\arraystretch}{1.2}
\setlength\tabcolsep{4.5pt}
\vspace{-1mm}
\resizebox{0.8\linewidth}{!}{
\begin{tabular}{llcc|ccccccc}
\toprule
\multicolumn{1}{l}{Setting} & \multicolumn{1}{l}{Model} & \#T-Param & Strategy & MME-P & DocVQA & InfoVQA &SQA-I& GQA & ChartQA & AI2D \\ 
\midrule
(1)&InternLM2                   & 1.8B   & Full   &   753    & 16.1   & 11.6  &  36.7   &  51.4   &   10.8     &  27.7    \\
(2) & (1) + V-Expert                      & 3.0B   & Full   &  948     &  18.6  &  11.9 &  37.7    &  53.0   & 11.1       & 26.6 \\
\rowcolor{gray!15}(3)&(1) + V-Expert                      & 1.2B   & Delta   &    995   &  18.9  & 14.6 & 56.5   &  53.4   &    13.5    & 42.7\vspace{-0.6mm} \\
\rowcolor{gray!15}(4)& (3) + A-Expert & 3.0B &Delta & 1000 &  21.1 & 15.2 & 57.8 &53.7 & 12.5 & 43.7\\
\bottomrule
\end{tabular}
}
\label{ablation_mmoe}
\end{table*}

\begin{figure*}[htbp]
    \centering
    \includegraphics[width=0.9\linewidth]{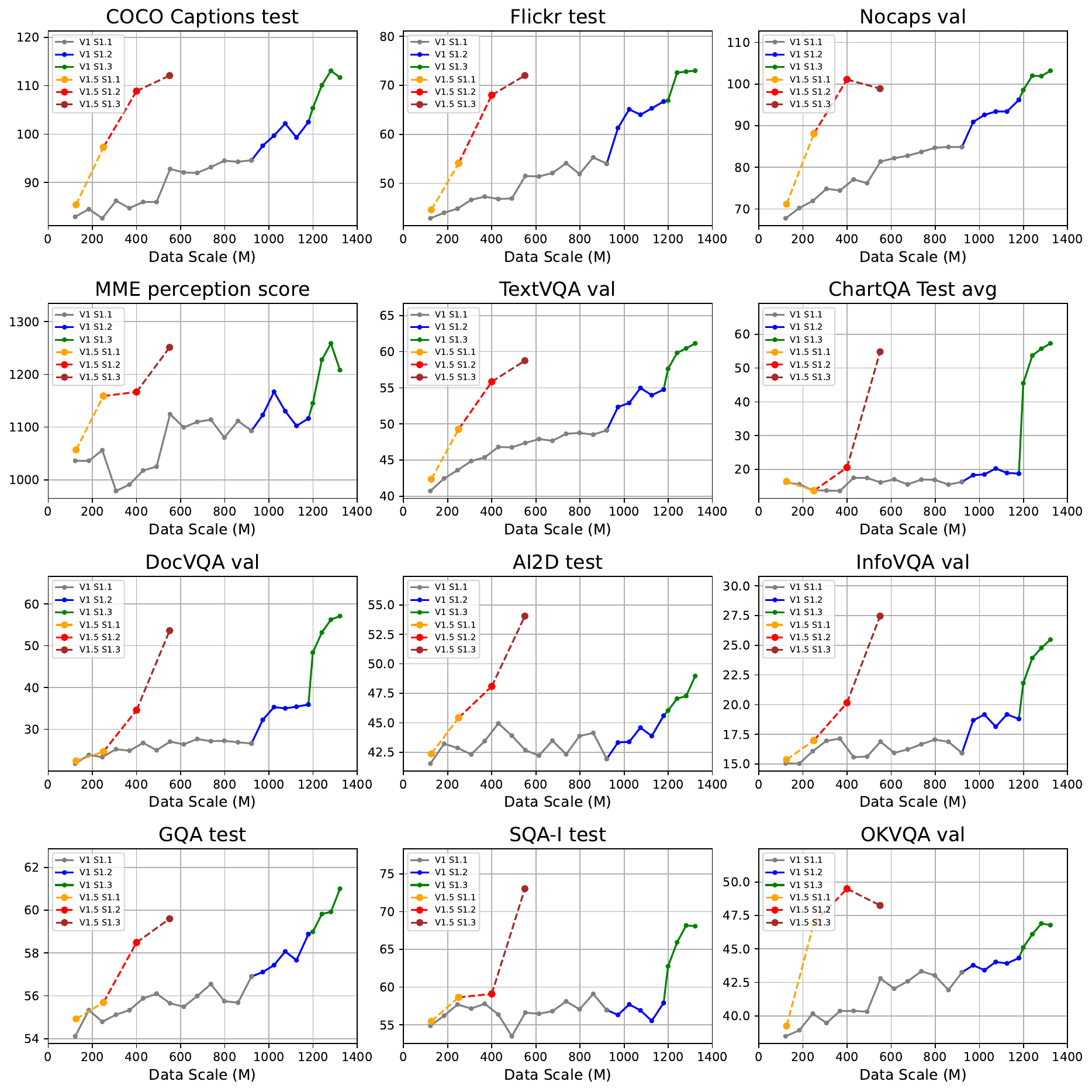}
    \caption{\textbf{Ablation studies of EViP  and EViP++ with the increase of pre-training data size across three sub-stages: (S1.1) Concept learning; (S1.2) Semantic learning; (S1.3) Alignment learning.} For each data point, we fine-tune the corresponding pre-trained model on the instruction data of LLaVA-665k and obtain the downstream performance.}
    \vspace{2mm}
    \label{fig:data_curve_full}
\end{figure*}

\begin{table}[t]
\small
\centering
\vspace{2mm}
\caption{\textbf{Ablation of freezing and unfreezing  attention in alignment learning.}  ``T-Param'' refers to trainable parameters. All models are pre-trained on 20 millions of data in alignment learning and fine-tuned on LLaVA-665k~\cite{VLM:LLaVA-1.5}.  Base model is the Mono-InternVL.} 
    % \vspace{-.5em}
\renewcommand{\arraystretch}{1.2}
\setlength\tabcolsep{1pt}
\vspace{-1mm}
\resizebox{\linewidth}{!}{
\begin{tabular}{lc|ccccccc}
\toprule
\multicolumn{1}{l}{Methods} & \#T-Param   & MME-P & DocVQA & InfoVQA &SQA-I& GQA & ChartQA & AI2D \\ 
\midrule
Freeze                   & 1.2B      &   1136   & 39.5   & 19.7  &  56.5   &  59.1   &   27.2     &  44.1    \\
\rowcolor{gray!15} Unfreeze                      & 1.8B      &  \textbf{1153}     &  \textbf{49.3 } &  \textbf{22.7} &  \textbf{61.8}    &  \textbf{59.9}   & \textbf{49.5}      & \textbf{46.4}  \\
\bottomrule
\end{tabular}
}
\label{ablation_attention}
\end{table}

\begin{table}[t]
\caption{\textbf{Ablation of S1.2 \textit{vs.} Longer training iterations of S1.1.} Both models are  Mono-InternVL and fine-tuned on LLaVA-665k.}
\renewcommand{\arraystretch}{1.4}
\setlength\tabcolsep{1pt}
\resizebox{1.0\columnwidth}{!}{
    \begin{tabular}{lcccccc}
    \toprule
     \textbf{Data}  & \textbf{TextVQA} & \textbf{SQA-I} & \textbf{GQA} &\textbf{DocVQA}  &\textbf{ChartQA} &\textbf{InfoVQA}\\
    \midrule
    1024M (S1.1)& 49.42 & 55.97 & 55.77 & 29.86 & 14.76 & 17.47 \\
     \rowcolor{gray!15} 922M (S1.1) + 102M (S1.2) & \textbf{52.90} & \textbf{57.71} &\textbf{ 57.43} & \textbf{35.32}  & \textbf{18.52} & \textbf{19.17} \\
    
    \bottomrule
    \end{tabular}
}
\label{ablation:longer_s11}
\end{table}

\begin{table}[t]
\caption{\textbf{Ablation of combining and separating S1.1 and S1.2.} Both models are   Mono-InternVL and fine-tuned on LLaVA-665k.}
\renewcommand{\arraystretch}{1.2}
\setlength\tabcolsep{1pt}
\resizebox{1.0\columnwidth}{!}{
    \begin{tabular}{lcccccc}
    \toprule
    \textbf{Strategy}  & \textbf{TextVQA} & \textbf{SQA-I} & \textbf{GQA} &\textbf{DocVQA}  &\textbf{ChartQA} &\textbf{InfoVQA}\\
    \midrule
    Combined (62M) & 38.61 & \textbf{56.92} & 54.72 &22.26   &14.36 &15.10 \\
    \rowcolor{gray!15}Separated (31M+31M) &\textbf{39.75}  &54.98  &\textbf{55.51}  &\textbf{22.29}  & \textbf{15.60}&\textbf{15.99}  \\
    
    \bottomrule
    \end{tabular}
}
\label{ablation:combine_s11_s12}
\end{table}

\begin{table}[t]
\centering
\caption{\textbf{NLP results of shared and unshared (\emph{i.e.} separated vision and text experts) architectures.} We use  Mono-InternVL as the base model, which is trained with 60M S1.1 data and then fine-tuned on LLaVA-665k.}
\renewcommand{\arraystretch}{1.2}
\setlength\tabcolsep{7pt}
\resizebox{1.\columnwidth}{!}{
    \scriptsize
    \begin{tabular}{lcccc}
    % \toprule
    \hline
    \textbf{Architecture} &  \textbf{MMLU} & \textbf{CMMLU} & \textbf{AGIEval} &\textbf{MATH} \\ 
    \hline
    % \midrule
    Shared                    &   26.17   & 25.40   & 14.89  &  0.24    \\
    \rowcolor{gray!15} Unshared  &   \textbf{42.30}   &  \textbf{41.11}  & \textbf{31.30}& \textbf{1.90}  \\
    % \bottomrule
    \hline
    \end{tabular}
}
\label{ablation:shared}
\end{table}

\begin{table}[t]
\centering
\caption{\textbf{Comparison of Mono-InternVL, Mono-InternVL-1.5 and existing monolithic MLLMs on four common NLP tasks. } Except for Chameleon, models are evaluated using opencompass toolkit~\cite{opencompass2023}. }
\label{NLP_TABLE}
\renewcommand{\arraystretch}{1.2}
\setlength\tabcolsep{1.5pt}
\resizebox{\linewidth}{!}{
\begin{tabular}{lc|ccccc}
\toprule
\multicolumn{1}{l}{\textbf{Models}} & \multicolumn{1}{l|}{\textbf{\#A-Param}} & \textbf{MMLU} & \textbf{CMMLU} & \textbf{AGIEval} & \textbf{MATH}  \\ \hline
InternLM2-Chat~\cite{cai2024internlm2}       & 1.8B                           & 47.1 &  46.1     & 38.8    &    13.9          \\ 
% Vicuna~\cite{TransF:Vicuna} & 7B & - & 37.4 & 26.6 & 2.76 \\ 
\hline
EVE~\cite{diao2024EVE}                        & 7B                             &   43.9   &   33.4    &     22.6    &     0.7        \\
Chameleon~\cite{team2024chameleon}                  & 7B                             & 52.1 & -     & -       & 11.5         \\
\rowcolor{gray!15} Mono-InternVL              & 2B                             & 45.1 &   44.0    & 40.9   & 12.3\vspace{-0.6mm}          \\ 
\rowcolor{gray!15} Mono-InternVL-1.5 & 2B & 44.7 & 41.7 &  38.9&  15.1\\\bottomrule
\end{tabular} }
\end{table}

\subsection{Main Results}

\begin{figure*}[t!]
    \centering
    \includegraphics[width=\linewidth]{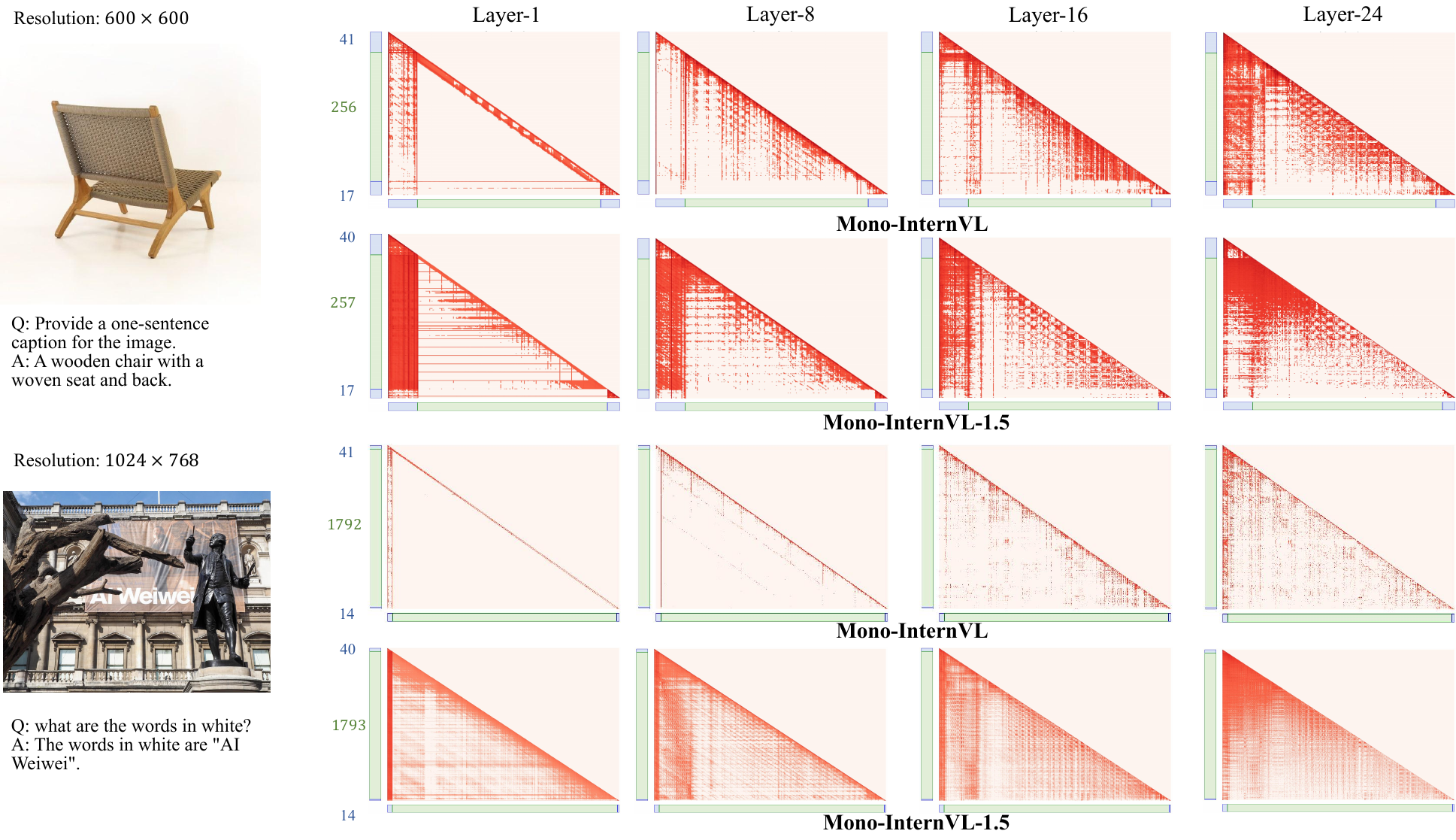}
    \vspace{-0.5em}
    \caption{\textbf{Visualization of attention maps in Mono-InternVL and Mono-InternVL-1.5.}  
     The first blue segment, green segment and the second green segment  in the axes represent the system prompt tokens (text), image tokens (visual) and user prompt tokens (text), respectively.   The numbers on the left side of  attention maps indicate the number of tokens.
    }
    % \vspace{-.5em}
    \label{vis:attention}
\end{figure*}

\textbf{Comparisons with modular MLLMs.} In Tab.~\ref{tab:vqa_benchmark} and ~\ref{tab:multimodal_benchmark}, we compare Mono-InternVL, Mono-InternVL-1.5 and existing MLLMs on 15 multimodal benchmarks.  The first observation is that  most modular MLLMs outperform existing monolithic MLLMs by significant margins. For example,  the average performance of  InternVL-1.5-2B~\cite{VLM:InternVL-1.5} on 9 MLLM benchmarks greatly exceeds the SoTA monolithic MLLM, \emph{i.e.,} + 15.5\% over EVE-7B (HD)~\cite{diao2024EVE}.  These results strongly suggest  the challenges  in  existing monolithic MLLMs.  In contrast,  Mono-InternVL-2B with a slightly smaller model size can even outperform the  modular baseline, \emph{i.e.,} + 0.8\% against  InternVL-1.5-2B on average.  Notably, Mono-InternVL-2B demonstrates distinct advantages on MathVista and OCRBench, suggesting its seamless text recognition and reasoning capabilities.  Compared to SoTA modular MLLM, \emph{i.e.,} Qwen2VL~\cite{Qwen2vl}, Mono-InternVL-2B  is still comparable in most benchmarks, \emph{e.g.,}   MathVista. We also observe that Mono-InternVL is still inferior to  InternVL-1.5 on high-resolution benchmarks, \emph{e.g.,} -12.4\% on InfoVQA. This may be because the relatively shallow model depth limits the visual encoding ability of Mono-InternVL, as shown in Fig.~\ref{vis:attention}.

\textbf{Comparisons with monolithic MLLMs.}  Compared to existing monolithic MLLMs, performance gains of Mono-InternVL become  distinct. For example, compared to  EVE-7B (HD)~\cite{diao2024EVE},   Mono-InternVL achieves up to 15.4\% average gains on   VQA tasks. Note that EVE-7B requires high-quality data for pre-training, so scaling it with more data remains challenging.  Furthermore, compared to Emu3~\cite{emu3},   Mono-InternVL still demonstrates better results on 9 of 12 benchmarks, while using much fewer parameters.  Compared to Mono-InternVL, Mono-InternVL-1.5 shows comparable or even better performance on multiple benchmarks. As shown in  Tab.~\ref{tab:vqa_benchmark}, on OCR-related benchmarks,  Mono-InternVL-1.5 outperforms Mono-InternVL by large margins, \emph{e.g.,} +1.7 on DocVQA and +4.9 on InfoVQA. On common MLLM benchmarks,  advantages of Mono-InternVL-1.5 are also obvious, \emph{e.g.,} +13.9\% on MMVet against Mono-InternVL.  Compared to the newly proposed monolithic MLLM called VoRA~\cite{vora},  Mono-InternVL-1.5  achieves better performance on most benchmarks, \emph{e.g.,} +15.0\% on textVQA. 
Compared to Mono-InternVL, Mono-InternVL-1.5 shows comparable or even better performance on multiple benchmarks. As shown in  Tab.~\ref{tab:vqa_benchmark}, on OCR-related benchmarks,  Mono-InternVL-1.5 outperforms Mono-InternVL by large margins, \emph{e.g.,} +1.7\% on DocVQA and +4.9\% on InfoVQA. On common MLLM benchmarks,  advantages of Mono-InternVL-1.5 are also obvious, \emph{e.g.,} +13.9\% on MMVet against Mono-InternVL.

In Tab.~\ref{NLP_TABLE}, we further compare the NLP ability of Mono-InternVL with existing monolithic MLLMs.  From this table, we notice that Mono-InternVL can well preserve its pre-trained NLP ability, which retains similar performance with InternLM2-Chat. However,   monolithic MLLMs like EVE, even with larger parameter size, are still inferior to Mono-InternVL in multiple NLP tasks.  In addition, we also find that Mono-InternVL-1.5 has a slight performance drop on some NLP tasks, but still outperforms EvE by margins. Considering the much cheaper training cost than Mono-InternVL, such performance drop is still acceptable.   These results further confirm the advantages of Mono-InternVL and Mono-InternVL-1.5  against existing monolithic MLLMs.

\textbf{Comparisons of pre-training results.} In Tab.~\ref{few_shot_result}, we further compare the pre-training performance of Mono-InternVL and existing MLLMs.  We observe that after concept and semantic learning, Mono-InternVL-S1.2 already exceeds existing modular MLLMs, \emph{e.g.,} +13.8 CIDEr over MM1~\cite{VLM:MM1} on COCO Captions, demonstrating that Mono-InternVL-S1.2 is effective in capturing basic multimodal relationships.  
When compared with monolithic MLLMs, Mono-InternVL also shows superior performance. For instance, even though Chameleon has a much larger model size, it is still inferior to Mono-InternVL-S1.3 by -2.6 CIDEr on Flickr30k~\cite{Datasets:Flickr30k}.  Compared to Mono-InternVL, Mono-InternVL-1.5 demonstrates superior training efficiency. With a total of 0.5 billion pre-training data, Mono-InternVL-1.5 reaches very competitive zero-shot performance against Mono-InternVL, \emph{e.g.,} 76.7 \textit{vs.} 77.3 on Flickr30k.  It is also  worth noting that pre-training in Mono-InternVL-1.5 only consumes about 0.5B image-text pairs, but the cost in MM1 and Flamingo is much more expensive, \emph{e.g.,} more than 2B data.   
These results further confirm  the effectiveness of EViP and EViP++.

\subsection{Ablation studies.}
 \textbf{Cumulative ablations of Mono-InternVL and Mono-InternVL-1.5.} To validate the design of  Mono-InternVL, we conduct extensive ablation studies. 
 Specifically, Tab.~\ref{ablation_mmoe} compares different strategies for visual pre-training.  The first row is the common strategy used in  existing monolithic MLLMs, \emph{i.e.,} full tuning of the LLM,  which yields the worst downstream performance in the table. After employing visual experts (the second row), such a full-tuning strategy becomes more effective, \emph{e.g.,} +1.6\% on GQA.  These comparisons well confirm the sub-optimal  design of the shared architecture for joint vision and language modeling.  We also observe that the delta tuning strategy greatly benefits the visual pre-training, providing +18.8\% and 16.1\% gains on SQA-I and AI2D, respectively.  Compared to full tuning,  delta tuning can effectively preserve the knowledge of the pre-trained LLM, which is also crucial for   multimodal modeling.   

\textbf{Impact of scaling data size in EViP and EViP++.}
Fig.~\ref{fig:data_curve_full} further demonstrates the relationship between downstream performance and pre-training data size.  We can observe that performance of  Mono-InternVL will gradually reach an upper bound in the concept learning.  Through additional semantic learning and alignment learning, capabilities of Mono-InternVL consistently boost as the data size increases.  It is important to note that that the alignment learning  plays a significant role for VQA tasks, which can provide sufficient task-related knowledge, \emph{e.g.,} OCR knowledge. These results suggest that the low-quality data of S1.1 contribute less to the performance than high-quality ones. Therefore, Mono-InternVL-1.5 adopts a new design principle for data organization, \emph{i.e.,} small in quantity but high in quality. As shown in Figure ~\ref{fig:data_curve_full}, Even with much less noisy data,  Mono-InternVL-1.5  can easily achieve similar performance to Mono-InternVL. Furthermore, thanks to the new architecture, the training efficiency of Mono-InternVL-1.5 greatly exceeds that of Mono-InternVL, further reducing the  requirement of data size. These results not only demonstrate  the data scalability of Mono-InternVL and Mono-InternVL-1.5, but also confirm  the  coarse-to-fine learning of EViP and the data efficiency of EViP++.

\begin{table}[t!]
\caption{
\textbf{Latency comparison of fused cuda kernel and PyTorch implementation for multimodal MoEs.}
The results are reported in $\mu$s.  ``Linear MoE'' and ``MLP MoE'' is used in self-attentions and feed-forward networks, respectively. Speed is tested on a single A100 GPU.
}
\resizebox{1.0\columnwidth}{!}{
\begin{tabular}{ccccccc}
\toprule 
\multirow{2}{*}{\textbf{Method}} & \multicolumn{6}{c}{\textbf{Sequence Length}}\tabularnewline
\cmidrule{2-7}
 & 2K & 4K & 16K & 32K & 64K & 128K\tabularnewline
\midrule
\rowcolor{gray!15}
\multicolumn{7}{c}{\textit{Linear MoE}\quad (2048$\rightarrow$4096)}\tabularnewline
\midrule
Pytorch & 508 & 846 & 3,169 & 6,317 & 12,520 & 49,497\tabularnewline
Fused Kernel & 276 & 436 & 1,769 & 2,745 & 5,408 & 21,317\tabularnewline
Speedup & \textbf{1.84}$\times$ & \textbf{1.94}$\times$ & \textbf{1.79}$\times$  & \textbf{2.30}$\times$ & \textbf{2.31}$\times$ &\textbf{2.32}$\times$ \\
\midrule
\rowcolor{gray!15}
\multicolumn{7}{c}{\textit{MLP MoE}\quad(2048$\rightarrow$
8192$\rightarrow$ 2048)}\tabularnewline
\midrule
Pytorch & 2,063 & 3,968 & 15,008 & 28,948 & 59,237 & 117,824\tabularnewline
Fused Kernel & 1,204 & 2,305 & 8,614 & 15,821 & 34,301 & 68,064\tabularnewline
Speedup & \textbf{1.71}$\times$ & \textbf{1.72}$\times$ & \textbf{1.74}$\times$  & \textbf{1.82}$\times$ & \textbf{1.72}$\times$ &\textbf{1.73}$\times$ \\
\bottomrule
\end{tabular}}
\label{tab:op-latency}
\end{table}

\begin{table*}[t]
\small
\centering
\renewcommand{\arraystretch}{1.2}
\caption{\textbf{Inference speed comparison of InternVL-1.5, Mono-InternVL,  and Mono-InternVL-1.5.} Models are deployed on an NVIDIA A100 GPU using LMDeploy with Pytorch backend~\cite{2023lmdeploy}. We use a concurrency of 16 and the number of output tokens fixed as 120. ``TTFT'' and ``TPS'' denotes the time to first token in seconds and throughput in tokens per second, respectively. 
}
\setlength\tabcolsep{10pt}
\resizebox{\linewidth}{!}{
\begin{tabular}{lccc|ll}
\toprule

\multirow{2}{*}{Model} & \#Image & \#Text & \#Total Input & \multirow{2}{*}{TTFT} & \multirow{2}{*}{TPS} \\
& Tokens & Tokens & Tokens & & \\
\hline
InternVL-1.5-2B  & 768 & 256 & 1024 & 0.242 & 382 \\
Mono-InternVL-2B & 768 & 256 & 1024 & 0.090 & 436 \\ %  (-63\%)  (+14.1\%)
Mono-InternVL-1.5-2B & 768 & 256 & 1024 & 0.092 & 433  \\
Mono-InternVL-1.5-2B + Fused Kernel & 768 & 256 & 1024 & 0.083 (-65.7\%) & 467 (+22.3\%) \\
\hline
InternVL-1.5-2B  & 1792 & 256 & 2048 & 0.453 & 183 \\
Mono-InternVL-2B & 1792 & 256 & 2048 & 0.151 & 232 \\ %  (-67\%)  (+27\%)
Mono-InternVL-1.5-2B  & 1792 & 256 & 2048 & 0.151 & 221 \\
Mono-InternVL-1.5-2B + Fused Kernel & 1792 & 256 & 2048 & 0.139 (-69.3\%) & 255 (+39.3\%) \\
\hline
InternVL-1.5-2B  & 3840 & 256 & 4096 & 1.938 & 52 \\
Mono-InternVL-2B & 3840 & 256 & 4096 & 0.795 & 68 \\ %  (-59\%)  (+31\%)
Mono-InternVL-1.5-2B & 3840 & 256 & 4096 & 0.810 & 61 \\
Mono-InternVL-1.5-2B + Fused Kernel & 3840 & 256 & 4096 & 0.659 (-66.0\%) & 77 (+48.1\%)\vspace{-0.6mm} \\ % 

\bottomrule
\end{tabular}
}
\label{tab:speed}
\end{table*}

\textbf{Ablations of micro-designs in Mono-InternVL.}
In Tab.~\ref{ablation_attention}, we examine the effects of freezing and unfreezing attention layers in alignment learning. We observe that unfreezing attention results in consistent improvements across all metrics, suggesting that it is crucial to optimize the multi-head attentions in this sub-stage for better vision-language alignment. To validate the effectiveness of synthetic data in S1.2, we compare two models: training S1.1 + S1.2 and training S1.1 only. Both models use the same amount of training data. From Tab.~\ref{ablation:longer_s11}, we observe that synthetic data helps to improve the performance. In Tab.~\ref{ablation:combine_s11_s12}, we examine whether we can merge S1.1 and S1.2 into one stage with a small amount of data, and find that separated stages have slight advantages. Finally, we further conduct experiments by removing vision experts and using shared FFN for vision and text, and evaluating its performance on NLP benchmarks. In Tab.~\ref{ablation:shared}, using shared architecture significantly affects the NLP performance, suggesting that it is necessary to use separate experts to preserve the pre-trained language capability.

\textbf{Ablations of the fused CUDA kernel.}
In Tab.~\ref{tab:op-latency}, we  compare the latency of fused CUDA kernel and PyTorch implementation. From this table, we observe that our fused CUDA kernel significantly outperforms the  PyTorch implementation. In the setting of "Linear MoE", the fused CUDA kernels achieve up to 2.32 times speedup against the  PyTorch implementation. Similar efficiency can also be  observed in the setting of ``MLP MoE'', \emph{e.g.,} up to 1.82 times speedup.  As the sequence length increases, the advantages of our fused CUDA kernel are consistent, which greatly confirm its technical contribution.

\textbf{Comparison of inference efficiency.} 
In Tab.~\ref{tab:speed}, we compare the inference speed of Mono-InternVL, Mono-InternVL-1.5 and InternVL-1.5 using the popular deployment library LMDeploy~\cite{2023lmdeploy}. From this table, we can find that due to the elimination of visual encoder, Mono-InternVL demonstrates superior efficiency under different number of input tokens. In particular, the first-token time  is greatly reduced in Mono-InternVL, \emph{e.g.,} up to -67\% against InternVL-1.5.  Benefiting from this, the overall throughput is correspondingly increased by around 31\%.  Compared to Mono-InternVL, the latency of Mono-InternVL-1.5 is slightly increased due to the additional visual experts in attentions.  After equipping with our fused CUDA kernel, we observe significant  improvements in inference efficiency, -19\% of TTFT against Pytorch implementation.  These results not only  validate the efficiency of Mono-InternVL and Mono-InternVL-1.5, but also confirm the benefit of our fused CUDA kernel.

\subsection{Visualizations} 
\textbf{Attention patterns of Mono-InternVL and Mono-InternVL-1.5.} To gain in-depth insights into Mono-InternVL and Mono-InternVL-1.5, we visualize its attention maps of different layers in Fig.~\ref{vis:attention}. From Fig.~\ref{vis:attention},  we can draw  two noteworthy conclusions.  Firstly, despite the global connectivity in the Transformer architecture, we find locality still exists in the visual encoding of shallow layers.  As shown in Fig.~\ref{vis:attention}, within the first layer, visual tokens only interact with their nearby content, resulting in patterns that are highly similar to those generated by convolutional neural networks~\cite{CNN:Resnet16}.  Second, modalities exhibit little interaction in shallow layers but gradually merge as the layers become deeper. The attention weights between visual and textual tokens are extremely low in the first layer and become higher in deeper layers.  In Mono-InternVL-1.5, the attention maps demonstrate a slightly different pattern. In particular,  after using visual experts in attentions, attention weights between modalities become larger. As shown in Fig.~\ref{vis:attention}, language tokens focus more densely on visual tokens, which confirms the advantage of visual experts in visual-language alignment.  We hope these examples will provide useful hints for the design of monolithic MLLMs.   

\section{Conclusion}
In this paper, we propose Mono-InternVL, a monolithic MLLM that integrates visual encoding and textual decoding  into a single LLM. In Mono-InternVL, a group of visual experts is embedded into the pre-trained LLM using a mixture-of-experts mechanism. By freezing the  LLM, Mono-InternVL ensures that visual capabilities are optimized without undermining the pre-trained language knowledge.  Then, an innovative Endogenous Visual Pre-training (EViP) is introduced to achieve coarse-to-fine visual learning of Mono-InternVL.  To further improve the  efficiency, we present EViP++ and propose a cheaper and faster model called Mono-InternVL-1.5.  Compared to Mono-InternVL, Mono-InternVL-1.5 benefits from additional visual attention experts, efficient data organization, and the multimodal MoE fused CUDA kernel. With these designs, Mono-InternVL-1.5 reduces the data requirement and inference latency by 58\% and 19\%, respectively,  while reaching better performance.  Extensive experiments  not only showcase the advantages of each design in Mono-InternVLs, but also verify their effectiveness and efficiency compared to existing MLLMs.  Our work significantly pushes the boundaries of monolithic MLLMs, offering new possibilities for the advancement of MLLMs.

\bibliographystyle{IEEEtran}
\bibliography{arxiv}

\end{document}